\documentclass[11pt]{article}
\usepackage{geometry}
\usepackage{coling2020}
\usepackage{times}
\usepackage{url}
\usepackage{latexsym}
\usepackage{microtype}
\usepackage{graphicx}
\usepackage{array}
\usepackage{booktabs}
\usepackage{multirow}
\usepackage{algorithm} 
\usepackage{algpseudocode} 
\usepackage{amssymb,amsmath}

\colingfinalcopy 

\title{CNRL at SemEval-2020 Task 5: Modelling Causal Reasoning in Language with Multi-Head Self-Attention Weights based Counterfactual Detection}

\author{Rajaswa Patil\textsuperscript{1,2} \and Veeky Baths\textsuperscript{1,3} \\
  \textsuperscript{1}Cognitive Neuroscience Lab, BITS Goa \\
  \textsuperscript{2}Department of Electrical \& Electronics Engineering \\
  \textsuperscript{3}Department of Biological Sciences \\
  BITS Pilani K. K. Birla Goa Campus, India \\
  {\tt \{f20170334, veeky\}@goa.bits-pilani.ac.in}}

\date{}

\begin{document}
\maketitle
\begin{abstract}
      In this paper, we describe an approach for modelling causal reasoning in natural language by detecting counterfactuals in text using multi-head self-attention weights. We use pre-trained transformer models to extract contextual embeddings and self-attention weights from the text. We show the use of convolutional layers to extract task-specific features from these self-attention weights. Further, we describe a fine-tuning approach with a common base model for knowledge sharing between the two closely related sub-tasks for counterfactual detection. We analyze and compare the performance of various transformer models in our experiments. Finally, we perform a qualitative analysis with the multi-head self-attention weights to interpret our models' dynamics.
\end{abstract}

\section{Introduction}
    Causal reasoning is a process of detecting cause-effect relationships and is increasingly being used in artificial intelligence for improving generalization and interpretability. Modelling causal reasoning in language involves detecting such cause-effect relationships from natural language texts. A cause-effect relationship can be modelled as: \emph{Event A causes Event B}. Counterfactuals describe events counter to facts and hence naturally involve common sense and causal reasoning. A counterfactual can be modelled as a cause-effect relationship of the form: \emph{Event A could have caused Event B (Event A did not occur)}. 

    SemEval-2020 Task-5 \cite{yang-2020-semeval-task5} consists of two independent sub-tasks: a binary-classification task for detecting counterfactual statements and a span-detection task for detecting antecedent (\emph{cause}) - consequent (\emph{effect}) spans of given counterfactual statements (Table~\ref{table1}). In this work, we use multi-head self-attention weights from pre-trained transformer models \cite{DBLP:journals/corr/VaswaniSPUJGKP17} to capture the semantic interactions between the tokens of given text with respect to causal relations. We use  a fine-tuning approach with a common base model for knowledge sharing between these two closely related sub-tasks. The code for this work is made publicly available as a GitHub repository.\footnote{\url{https://github.com/rajaswa/counterfactual-detection-semeval-2020}}

\section{Background}
    Early work on causal reasoning and related tasks in natural language was based on various statistical and linguistic approaches \cite{DBLP:journals/corr/Asghar16a}. Recent work for causal reasoning related tasks involves deep learning based approaches. Causal reasoning can be achieved through extraction of cause-effect relations with CRF and LSTM based sequence labelling tasks \cite{dasgupta-etal-2018-automatic-extraction}. Counterfactuals can contain implicit causal relations (Table~\ref{table1}). Using multi-head self-attention at word level can help capture such implicit causal relations effectively \cite{liang2019multilevel}. Current benchmarks for modelling causal reasoning involves question-answering tasks \cite{gordon-etal-2012-semeval}. Using pre-trained transformer models have been effective on such tasks \cite{DBLP:journals/corr/abs-1904-09728}. Self-attention weights from such transformer models are usually structured in a 3-dimensional matrix. Using convolutional neural networks with  these self-attention weight matrices can be helpful for extracting semantic features for downstream NLP tasks \cite{10.1371/journal.pone.0222713}. Similar approaches can be used to extract features for detecting counterfactuals in natural language texts. Further, the multi-head self-attention attention weights from these models can also be used in interpretative qualitative analysis \cite{DBLP:journals/corr/abs-1905-09418}.

\begin{table*}[t]
\centering
\renewcommand\arraystretch{2}
\begin{small}
\begin{tabular}[t]{>{\centering}m{0.3\linewidth}>{\centering}m{0.3\linewidth}>{\centering\arraybackslash}m{0.3\linewidth}}
\toprule
\textbf{Counterfactual Statement} & \textbf{Antecedent} & \textbf{Consequent} \\
\midrule
"If I had 10 pharmacists who worked with me, I could reach 100 people more effectively." & \emph{If I had 10 pharmacists who worked with me} & \emph {I could reach 100 people more effectively} \\
\midrule
"Thanks for the article on this new term that fits me so well, wish all your articles were worthy of praise." & \emph{wish all your articles were worthy of praise} & - \\
\bottomrule
\end{tabular}
\end{small}
\caption{Example counterfactual statements from the task dataset}
\label{table1}
\end{table*}

\section{Methodology}

\subsection{Base Architecture}
\begin{figure}[h]
  \center
  \includegraphics[scale=0.75]{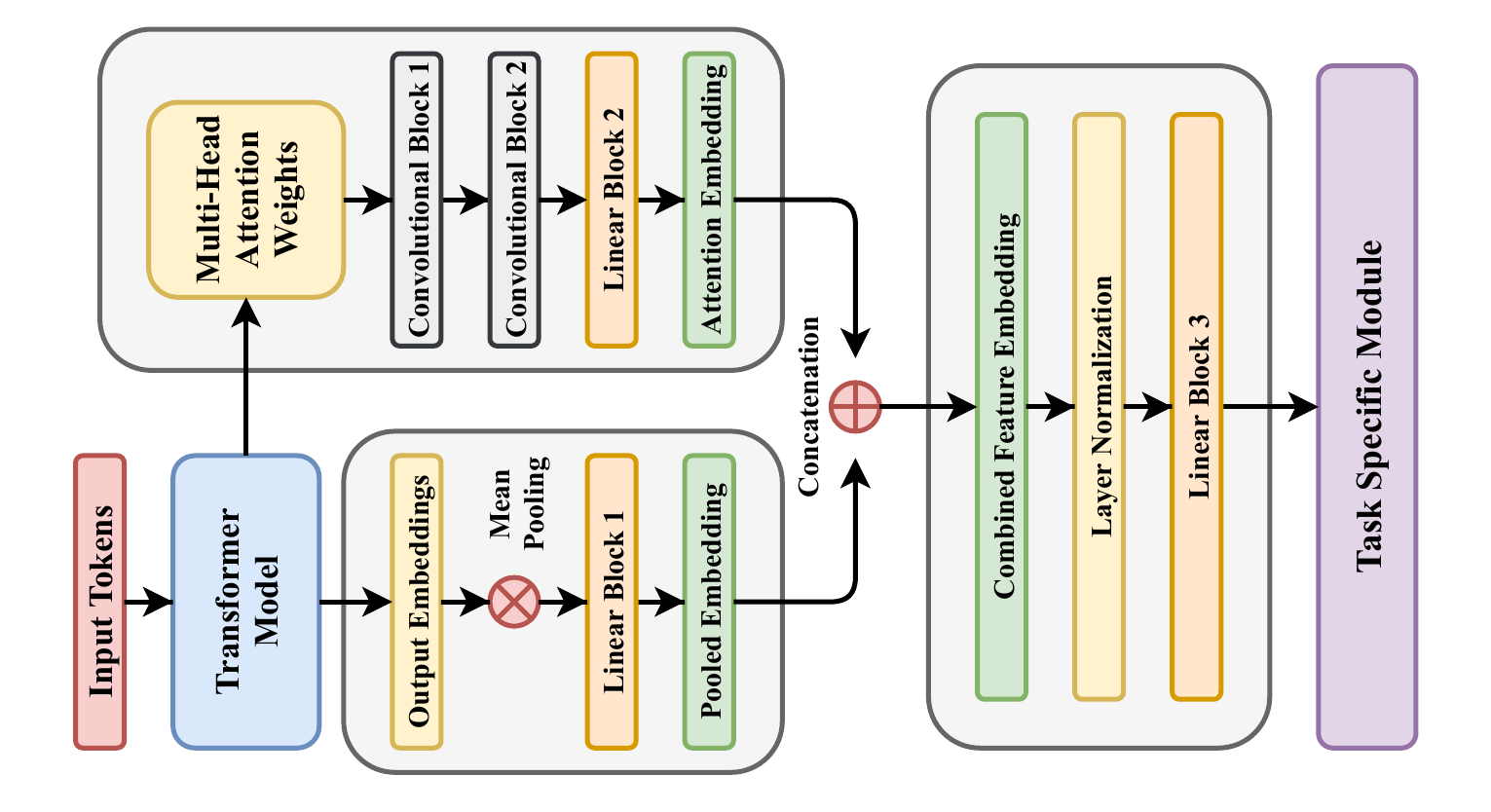}
  \caption{Base Architecture}
  \label{figure1}
\end{figure}

    We use the knowledge learned during the binary-classification counterfactual detection sub-task for the antecedent-consequent span-detection sub-task by defining a base architecture, common to both the sub-tasks (Figure~\ref{figure1}). The base architecture is used to extract task-specific features, which are further passed on to task-specific modules. We first train the base architecture with a binary-classification module for the first sub-task. Then, we replace the binary-classification module with a regression-module and fine-tune the already trained base architecture for the more complex second sub-task.

    For the base architecture, we use pre-trained transformer models to extract contextual output embeddings and multi-head self-attention weights from the tokenized input text. The output embeddings are passed through a pooling layer to get a pooled embedding. The multi-head self-attention weights are structured in a 3-dimensional matrix with the following dimensions: \emph{(number of attention heads, number of tokens in the text, number of tokens in the text)}. This matrix is passed through convolutional and linear blocks to get an attention embedding. The pooled embedding and the attention embedding are concatenated together to form a combined feature embedding. We apply a layer normalization operation on the combined feature embedding for better generalization and stability for knowledge sharing across the sub-tasks. The feature embedding is then passed through a linear block and fed to the task-specific module. A linear block is composed of a fully connected layer with ReLU activation and dropout regularization, and a convolutional block is composed of a 2D-convolution layer with batch normalization, ReLU activation and dropout regularization. Here, we experiment with various pre-trained transformer models which differ from each other in terms of pre-training approach, architecture and number of parameters: BERT \cite{devlin-etal-2019-bert}, RoBERTa \cite{DBLP:journals/corr/abs-1907-11692} (robust pre-training), XLNet \cite{DBLP:journals/corr/abs-1906-08237} (autoregressive model) and DistilBERT \cite{Sanh2019DistilBERTAD} (distilled model). Usually, the last layer of any transformer model gets quickly biased for any individually trained task. Here, we concatenate the output embeddings and multi-head self-attention weights from the last three layers of the transformer model so that more generalized features are learned by the common base architecture while training for each of the sub-tasks separately.

\subsection{Task Specific Modules}
    For the counterfactual detection sub-task, we have binary labels for various statements to be \emph{counterfactuals / non-counterfactuals}. We use a linear block with sigmoid activation as a binary-classification module for this task. For the second sub-task, we have character-level span locations (\emph{start-id} and \emph{end-id}) for the antecedent and consequent spans of the given counterfactual statements. This can be treated as a regression problem with 4 feature values. We use an another linear block with ReLU activation and 4 output neurons as a regression module for this task. The lengths of various counterfactual statements in the second sub-task vary considerably across the dataset. This induces a certain variance in the character-level span location features. To handle this variance, we scale each of these 4 span features by the length of the counterfactual statement. The span features are scaled down by the lengths of their respective counterfactual statements during training. Consequently, we scale up the predicted span features during inference to obtain the actual antecedent-consequent span locations. The functioning of the regression module for antecedent-consequent detection during training and inference is explained in Algorithm~\ref{algorithm1}.

\begin{algorithm}
	\caption{Counterfactual Antecedent-Consequent Detection} 
	\begin{algorithmic}[1]
		\While {Training}
		    \ForAll{$counterfactual\in\mathit{data}$}
		       \State $spans \Leftarrow (antecedent\_start, \ antecedent\_end, \ consequent\_start, \ consequent\_end)$
		       \State $spans \Leftarrow \frac{spans}{length(counterfactual)}$
		       \State $output \Leftarrow \textbf{RegressionModule}(counterfactual)$
		       \State $loss \Leftarrow \textbf{SmoothL1Loss}(output, \ spans)$
		    \EndFor
        \EndWhile
        
        \While {Inference}
		    \ForAll{$counterfactual\in\mathit{data}$}
		       \State $output \Leftarrow \textbf{RegressionModule}(counterfactual)$
		       \State $spans \Leftarrow output * length(counterfactual)$
		    \EndFor
        \EndWhile
        
	\end{algorithmic}
\label{algorithm1}
\end{algorithm}

\section{Experiments}
    For all our experiments, we use Binary Cross Entropy loss (\emph{for counterfactual detection}) and Smooth L1 loss (\emph{for antecedent-consequent span regression}) with Adam optimizer (\emph{with weight decay}) to train our models. We use the PyTorch implementations of the smallest base variants of pre-trained transformer models by Hugging Face\footnote{\url{https://huggingface.co/}} \cite{Wolf2019HuggingFacesTS} in our base architecture. We use a 90-10 data split for training and development purposes respectively. The data distribution across the splits is shown in Table~\ref{data-split}. We validate our models on F1 score (\emph{for counterfactual detection}) and Smooth L1 loss (\emph{for antecedent-consequent detection}). The evaluation metrics for the counterfactual detection task are the binary precision, recall and F1 scores. For the antecedent-consequent span detection task, the precision, recall and F1 score are defined as sequence labelling metrics with respect to the overlap between the predicted and the ground truth spans\footnote{\url{https://github.com/arielsho/Task-5_Baseline/blob/master/subtask2_baseline.py}}.

\begin{table}[h]
\centering
\resizebox{0.65\textwidth}{!}{%
\begin{tabular}{@{}cccc@{}}
\toprule
\textbf{Task}                                  & \textbf{Training} & \textbf{Development} & \textbf{Test} \\ \midrule
Counterfactual Detection & 11700    & 1300       & 7000 \\
Antecedent-Consequent Detection         & 3195     & 356        & 1950 \\ \bottomrule
\end{tabular}%
}
\caption{Data distribution across the splits (number of samples)}
\label{data-split}
\end{table}

\section{Discussion}
    For the final submission, we use RoBERTa and BERT as the transformer model in our base architecture for sub-task 1 and sub-task 2 respectively. On the final task leaderboard, our system ranks 13\textsuperscript{th} (0.845 F1) for the counterfactual detection sub-task and 7\textsuperscript{th} (0.688 F1) for the antecedent-consequent detection sub-task. Since we treat antecedent-consequent detection as a regression task, we do not monitor the \emph{Exact Match score} between the predicted and ground truth spans, which is more significant for token-level sequence labelling based approaches. We analyse the performance of various transformer models with hyperparameter tuning post evaluation (Table~\ref{results}). RoBERTa gives the best results for the counterfactual detection sub-task. Whereas, BERT gives the best results for the antecedent-consequent detection sub-task. DistilBERT, a considerably smaller model (65M parameters) shows comparable performance with the rest of the transformer models (110M+ parameters) for both the sub-tasks.

\begin{table}[h]
\centering
\resizebox{0.75\textwidth}{!}{%
\begin{tabular}{@{}ccccccc@{}}
\toprule
\multirow{2}{*}{\textbf{Transformer Model}} & \multicolumn{3}{c}{\textbf{Sub-Task 1}} & \multicolumn{3}{c}{\textbf{Sub-Task 2}} \\ \cmidrule(l){2-7} 
           & \textbf{F1} & \textbf{Recall} & \textbf{Precision} & \textbf{F1} & \textbf{Recall} & \textbf{Precision} \\ \midrule
BERT       &  0.824  &    0.787    &     0.863      &  \textbf{0.688}  &    \textbf{0.672}    &     0.74      \\ \midrule
RoBERTa    &  \textbf{0.863}  &    0.847    &    \textbf{0.879}       &  0.644  &   0.567     &     \textbf{0.822}      \\ \midrule
XLNet      &  0.823  &    \textbf{0.862}    &     0.788      & 0.634   &    0.554    &     0.816      \\ \midrule
DistilBERT &  0.788  &    0.825    &     0.754      &  0.639  &    0.566    &    0.812       \\ \bottomrule
\end{tabular}%
}
\caption{Post evaluation analysis of various transformer models in base architecture}
\label{results}
\end{table}

     Since BERT performs marginally better than rest of the transformer models for antecedent-consequent detection sub-task, we consider BERT for the further qualitative analysis. We inspect the multi-head self-attention weights from the final layer of BERT \cite{vig2019transformervis} to interpret the model's dynamics. Overall, the model assigns more attention to certain parts of text which are related to the conditional nature of the counterfactual statements. Moreover, we see that some of the attention-heads learn to assign more attention to some specific parts of the text. Head\textsuperscript{1,6} assign maximum attention to punctuation and Head\textsuperscript{2,4,12} focus more on the auxiliary verbs. Head\textsuperscript{3,11} attend to conjunctions and verbs which act as causal connectives in the text. Head\textsuperscript{5} and  Head\textsuperscript{7} attend to entities and numerical values respectively (if present). This property of linguistically selective-attention of the attention-heads can be observed in the following examples of antecedent-consequent spans detected by our system (rounded off to include partially covered words). Where, an underline indicates the detected antecedent and the detected consequent is made bold.\\
     
\begin{enumerate}

  \item \underline{If\textsuperscript{3,11} only\textsuperscript{3} Trump had listened to Chris\textsuperscript{5} Christie\textsuperscript{5}} \textbf{he wouldn't\textsuperscript{2,4,12} be in this mess.}
  
  \item \underline{If\textsuperscript{3,11} this were an open seat\textsuperscript{5}, you} \textbf{would\textsuperscript{2,4,12} have six, eight\textsuperscript{7}, maybe 12\textsuperscript{7} people running.}
  
  \item \underline{'I wish\textsuperscript{2,4,11} I was 40\textsuperscript{7} years old, but I'm not,' he} told\textsuperscript{12} POLITICO\textsuperscript{5}.
  
  \item \underline{I could\textsuperscript{2,4,11} have been you and you could\textsuperscript{3,12} have} been me.
  
  \item Of course \textbf{the company wouldn't\textsuperscript{2,4,7,12} have\textsuperscript{3} had to sell such a prized} asset if\textsuperscript{11} it \underline{had other options to raise\textsuperscript{5} capital.}
  
\end{enumerate}

\begin{figure}[h]
  \center
  \includegraphics[width=\textwidth]{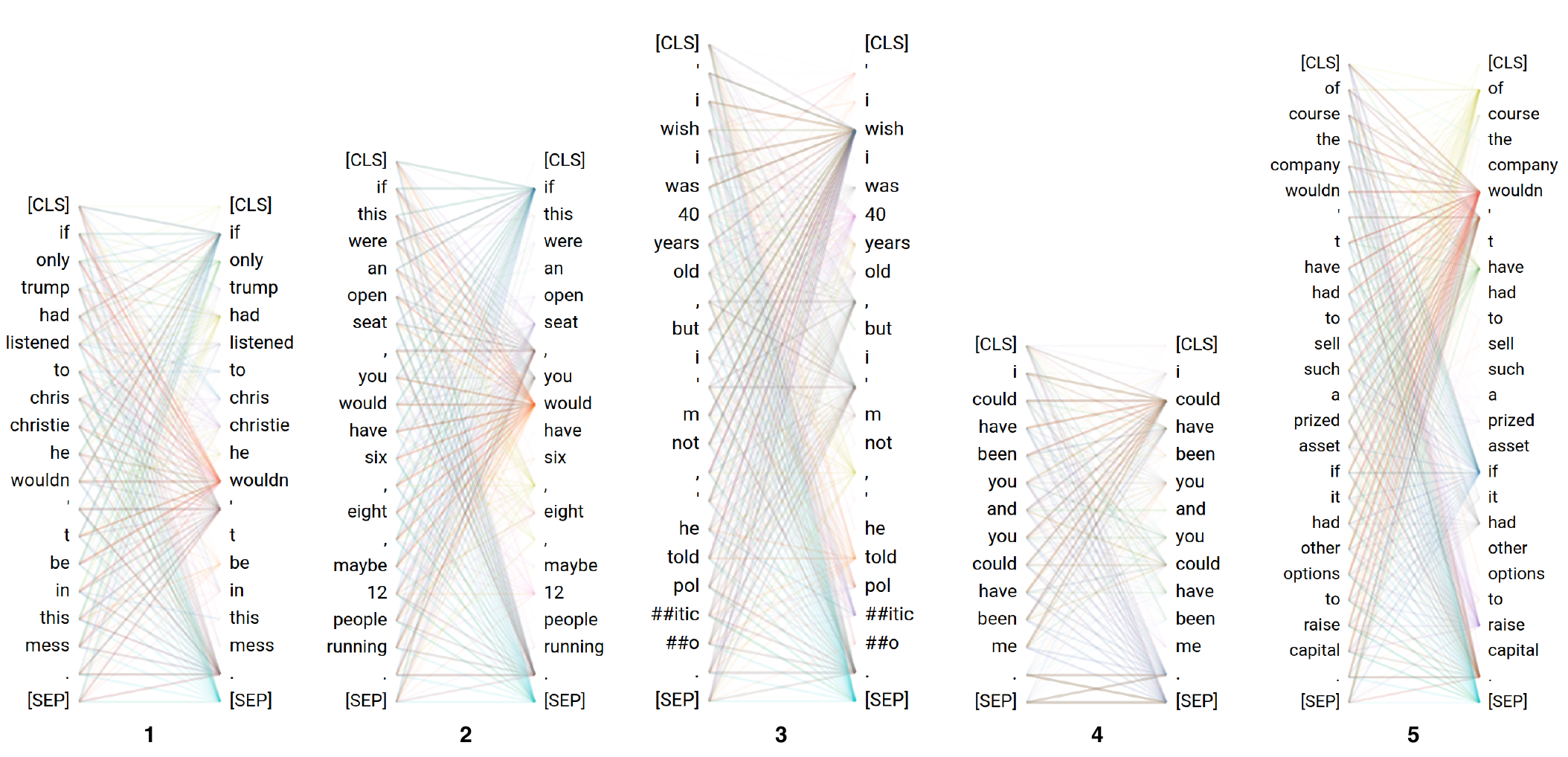}
  \caption{Head-wise self-attention weights visualization for BERT in base architecture}
  \label{figure2}
\end{figure}

The superscripts here represent the most attending attention-head(s) for the corresponding word. The same can be confirmed by a visualization (Figure~\ref{figure2}) of the head-wise color coded self-attention weights. For example 3 and 4, we have no consequent part in the text. Our system detects (0,0) as consequent span start and end locations for such counterfactual statements, indicating the absence of the consequent. The conjunctions (\emph{but\ /\ and}) in such counterfactual statements are ignored by the attention-heads. But the conjunctions (\emph{if}) in counterfactual statements with a consequent part (Example 1,2 and 5) are highly attended by the attention-heads through the tokens from entire sequence. This shows the ability of the model to differentiate the causal connectives from the non-causal ones in the text. Punctuation play an important role here as they are usually present near the boundaries of antecedent-consequent spans (Example 2 and 3). Auxiliary verbs (would, wouldn't, could, have) are assigned maximum attention across all the examples as they directly correspond to the conditional nature of counterfactual statements.

\section{Conclusion}
Our proposed approach uses multi-head self-attention weights from transformer models to detect causal relations for counterfactual detection in text. Through our experiments, we find that RoBERTa overall shows the best performance for counterfactual detection task and BERT performs the best for the antecedent-consequent detection task. We show that even smaller transformer models like DistilBERT perform counterfactual detection tasks effectively. With knowledge sharing between the two sub-tasks, our system detects antecedent-consequent spans in counterfactual statements with good efficiency by a simple regression over the spans. This can possibly be further improved by post-inference processing on the predicted spans or replacing the regression module with a token-level sequence labelling module. Further, we show that through our approach, the attention-heads attain a property of assigning linguistically selective-attention with respect to the conditional nature of the counterfactual statements.

\section*{Acknowledgements}
This work was carried out at the Cognitive Neuroscience Lab at BITS Goa\footnote{\url{http://bitscogneuro.com/}} through the funding from DST-CSRI – SR/CSRI/50/2014(G).

\bibliographystyle{coling}
\bibliography{semeval2020}

\end{document}